\documentclass{article}

\usepackage{microtype} 
\usepackage{graphicx} 
\usepackage{subfigure} 
\usepackage{graphicx} 
\usepackage{amsmath} 
\usepackage{amssymb} 
\usepackage{changepage} 
\usepackage[round]{natbib} 

\usepackage{xcolor}
\newcommand\hl[1]{%
  \bgroup
  \hskip0pt\color{red!80!black}%
  #1%
  \egroup
}

\usepackage{geometry}
\DeclareMathOperator*{\argmax}{arg\,max \text{ }} 
\DeclareMathOperator*{\argmin}{arg\,min \text{ }} 
\newcommand{\norm}[1]{\left\lVert#1\right\rVert} 
\newcommand{\dataset}[1]{\mathbf{#1}}

\title{A Linear Systems Theory of Normalizing Flows}
\author{
  Reuben Feinman\\
  New York University\\
  \texttt{reuben.feinman@nyu.edu}
  \and
  Nikhil Parthasarathy\\
  New York University\\
  \texttt{np1742@nyu.edu}
}
\date{}

\begin{document}
\maketitle

\begin{abstract}

Normalizing Flows are a promising new class of algorithms for unsupervised learning based on maximum likelihood optimization with change of variables. 
They offer to learn a factorized component representation for complex nonlinear data and, simultaneously, yield a density function that can evaluate likelihoods and generate samples.
Despite these diverse offerings, applications of Normalizing Flows have focused primarily on sampling and likelihoods, with little emphasis placed on feature representation.
A lack of theoretical foundation has left many open questions about how to interpret and apply the learned components of the model.
We provide a new theoretical perspective of Normalizing Flows using the lens of linear systems theory, showing that optimal flows learn to represent the local covariance at each region of input space. 
%
%
%
Using this insight, we develop a new algorithm to extract interpretable component representations from the learned model, where components correspond to Cartesian dimensions and are scaled according to their manifold significance.
In addition, we highlight a stability concern for the learning algorithm that was previously unaddressed, providing a theoretically-grounded solution to mediate the problem.
Experiments with toy manifold learning datasets, as well as the MNIST image dataset, provide convincing support for our theory and tools.
\end{abstract}

\section{Introduction}
\label{sec:intro}
A fundamental problem in unsupervised learning is to estimate the probability density function that underlies observations from of an unknown multidimensional data distribution. 
A faithful density estimate should, in theory, provide a handful of useful features. 
First, it provides a function to evaluate likelihoods for new observations that may or may not come from the training distribution. 
Second, it provides a procedure to generate samples from the distribution, mimicking the true causal generative process. 
Third, it provides a representation of the nonlinear degrees of freedom---i.e. the \textit{nonlinear components}---in the data manifold.

Normalizing Flows (NFs) \citep{Dinh2015, Rezende2015, Dinh2017, Kingma2018} are a promising new class of nonparametric density estimators that advertise all three of these desirable features within a single model. 
Given observations of a random variable (r.v.) $X$ with unknown distribution $p_X(x)$, the NF objective is to create an auxiliary r.v. $Y = f(X)$ such that $p_Y(y)$ has a simple, fully-factorized distribution. 
The change of variables theorem of calculus dictates that, for an invertible function $f: \mathcal{R}^D \rightarrow \mathcal{R}^D$ that maps a r.v. $X$ to a new r.v. $Y=f(X)$,  the densities $p_X(x)$ and $p_Y(y)$ are related by the formula
\begin{align}
    p_X(x) = p_Y(f(x))*\big| \text{det } J_f(x) \big|,
    \label{eq:change_of_variables}
\end{align}
where $J_f(x)$ is the Jacobian matrix of $f$ evaluated at $x$.
Normalizing Flows build on this relationship, capitalizing on the capacity of neural networks to approximate complex functions. 
They propose to use gradient-based optimization to estimate the function $f$ that maps observed r.v. $X$ to a new r.v. $Y$ with a target distribution that is fully-factorized.
Given target density $p_Y(y)$, the approach is to specify an invertible, parameterized function $f$ and to optimize this function such that the average log-likelihood of the observed data is maximized. 
For a dataset $\dataset{X} = \{x_1, ..., x_N\}$ containing $N$ observations, the objective is written
\begin{align}
    f^* &= \argmax_f \frac{1}{N} \sum_{n=1}^N \text{log }p_X(x_n) \nonumber \\
    &= \argmax_f \frac{1}{N} \sum_{n=1}^N \text{log }p_Y(f(x_n)) + \text{log } \big| \text{det } J_f(x_n) \big|.
    \label{eq:nice_objective}
\end{align}

This new learning framework offers a combination of useful features that have been pursued independently in prior literature (Sec. \ref{sec:related_work}), and it has been demonstrated successfully with raw, high-dimensional data such as natural images \citep[e.g.,][]{Dinh2017, Kingma2018, Hoogeboom2019}.
Nevertheless, progress in Normalizing Flows has been inhibited by a lack of theoretical foundation.
Applications of these models have focused primarily on sampling and likelihood computation, leaving many open questions about the learned component representations.
The default latent space of these models is whitened, meaning that insignificant components are indistinguishable from their significant counterparts.
Furthermore, the component axes in this space need not correspond to the Cartesian dimensions, making them difficult to identify and scale appropriately. 
To obtain a meaningful feature representation from these models, a better theoretical understanding is needed.

Our principal contribution is a new theoretical perspective of Normalizing Flows based on linear systems theory. 
We show that optimal flows learn to represent the local data covariance at each region of input space, and that properties of this covariance can be extracted via spectral decomposition of the Jacobian matrix (Sec. \ref{sec:background}).
Using this insight, we develop a new algorithm to compute un-whitened component representations with the learned model (Sec. \ref{sec:projections}).
Unlike the default component space, this new feature space provides meaningful distance measurements, and it facilitates visualization and interpretation of the learned components.
In addition to our feature extraction algorithm, we highlight a concern regarding the stability of the NF objective and offer a new regularization technique that stabilizes the learning algorithm (Sec. \ref{sec:tik_reg}).
To the best of our knowledge, we are the first to identify this concern explicitly, or to provide a theoretically-grounded solution.
We demonstrate these new tools on a handful of nonlinear data manifolds, as well as on the MNIST image dataset (Sec. \ref{sec:experiments}). Our regularized NF models produce robust component estimates that are consistent with the underlying distributions and are scaled according to manifold significance.

\section{Related Work}
\label{sec:related_work}
Nonparametric density estimation is a topic with a rich history in machine learning, and prior techniques have each focused on a subset of the desirable features we emphasize. 
Deep generative models such as Variational Autoencoders (VAEs) \citep{Kingma2013} and Generative Adversarial Networks (GANs) \citep{Goodfellow2014} can approximate the causal sampling procedure for unknown distributions, and they produce convincing samples of complex high-dimensional variables in many cases. 
However, the diversity of generated samples is difficult to measure, and the coherency of samples alone provides insufficient evidence for a generalized density estimate.
VAEs can also compute pseudo-likelihoods for new inputs; however, the metric provides only a lower bound on the true likelihood, and computations of this metric are approximate. 
Although both VAEs and GANs provide a mapping from independent components to complex distributions, neither has been carefully studied from the perspective of fundamental component analysis.

A separate class of algorithms emphasizes nonlinear component representations, using theoretically-grounded techniques based on spectral decomposition of similarity matrices and graphs \citep[e.g.,][]{Scholkopf1998, Tenenbaum2000, Roweis2000}. 
Spectral techniques can effectively estimate the independent components of nonlinear data manifolds in many cases; however, these algorithms do not provide a density model of the observed variable, and therefore they cannot draw likelihoods or evaluate samples. 
Furthermore, the similarity metrics used by these algorithms rely on measurements in Euclidean space, explaining perhaps why these techniques have had little success with raw, high-dimensional data like natural images.
With data of this kind, measurements in Euclidean space are ineffective at capturing perceptually-relevant similarities between observations \citep{Laparra2016, Wang2004}.

Both deep generative models and spectral techniques suffer from an important limitation as nonparametric estimators. Each of these algorithms requires \textit{a piori} knowledge of the intrinsic dimensionality in the data. 
This hyperparameter is specified through the bottleneck size of a VAE, the seed dimensionality in a GAN, or the number of components in spectral techniques.
Although cross-validation may be feasible in some cases, a more general estimator is desired that learns the intrinsic dimensionality of the data directly from observations.

We build on recent developments from Normalizing Flows, providing a new theoretical foundation that connects the NF learning algorithm to linear systems theory and fundamental component analysis.
We show that, with a new set of tools for regularization and component extraction, Normalizing Flows offer a unified framework for end-to-end density estimation, producing likelihoods, samples and components while requiring minimal a priori knowledge.

\section{Theory of Normalizing Flows}
\label{sec:background}
\begin{figure}[t]
    \centering
    \subfigure[Linear whitening (PCA)]{
        \centering
        \includegraphics[width=0.48\textwidth]{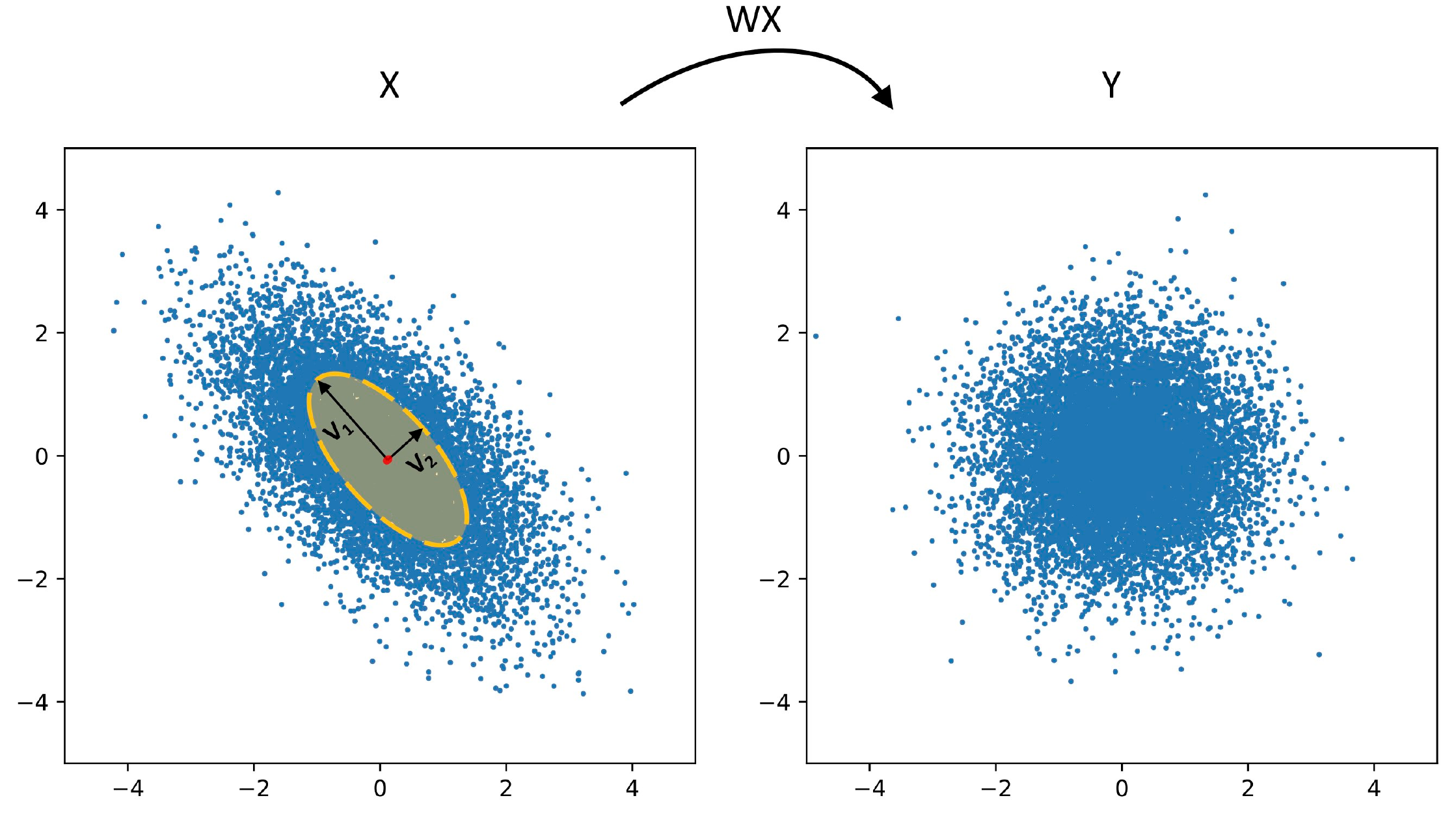}
    }
    \subfigure[Nonlinear whitening]{
        \centering
        \includegraphics[width=0.48\textwidth]{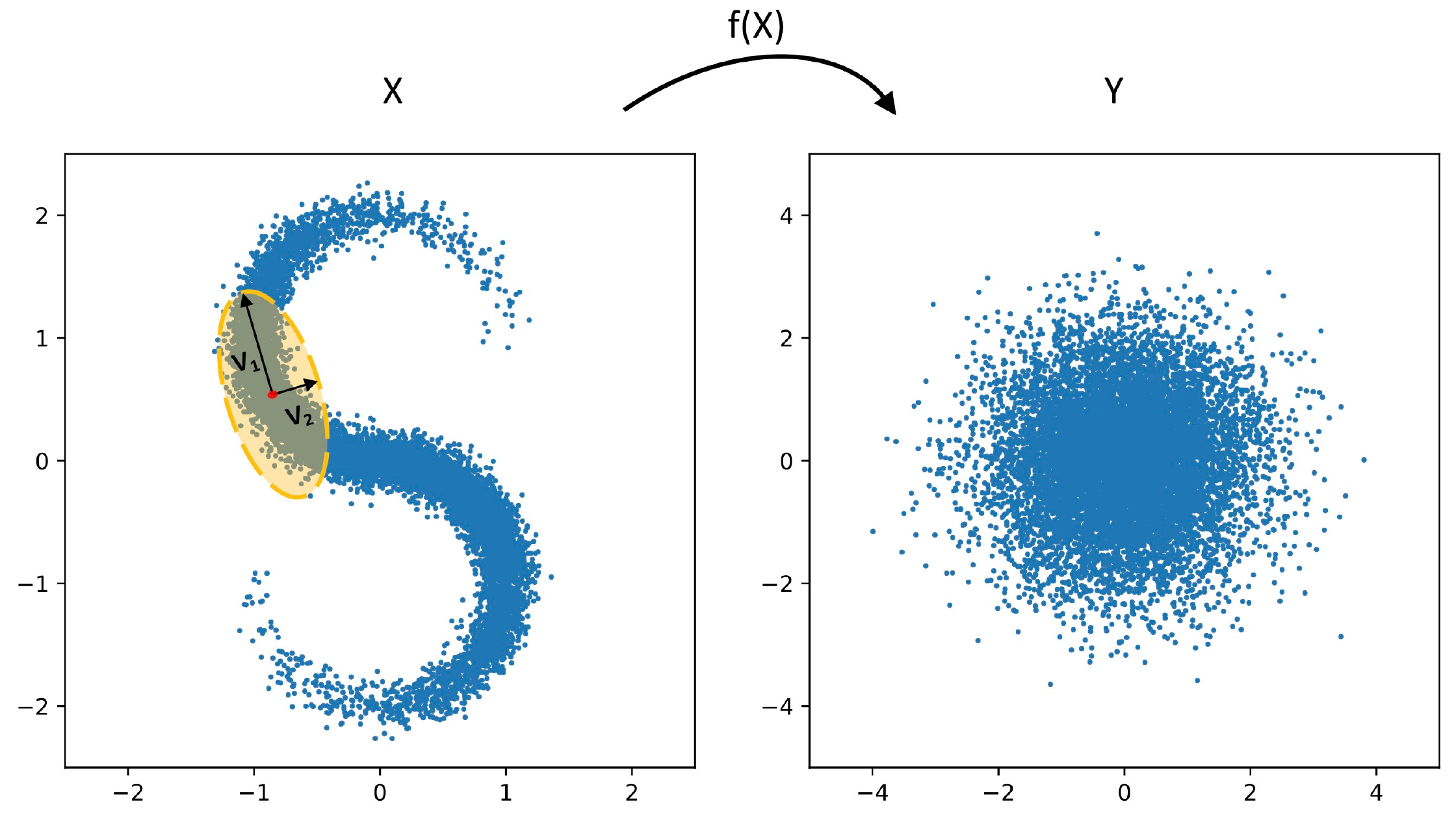}
    }
    \caption{Linear and nonlinear whitening transformations. (a) A linear whitening transformation $Y = WX$ maps a Gaussian r.v. $X \sim \mathcal{N}(0, \Sigma_X)$ to isotropic Gaussian $Y \sim \mathcal{N}(0,I)$. The covariance estimate, $\hat{\Sigma}_X = \big( W^TW \big)^{-1}$, is fixed across the input space. (b) A nonlinear whitening transformation $Y = f(X)$ maps a non-Gaussian r.v. $X \sim p_X(x)$ to isotropic Gaussian $Y \sim \mathcal{N}(0,I)$. The local covariance estimate, $\hat{\Sigma}(x) = \big(J_f(x)^T J_f(x)\big)^{-1}$, varies as a function of the input and is specified by the Jacobian matrix $J_f$. In both (a) and (b), blue dots show samples of the observed variable $X$ (left) and of the transformed variable $Y$ (right). Black vectors (left) convey the local component directions and magnitudes of $p_X$ at a particular input location.}
    \label{fig:linear_nonlinear_CA}
\end{figure}

We begin with a rigorous theoretical analysis of the Normalizing Flow objective. 
Throughout the section, assume we are given a dataset $\dataset{X} = \{x_1, ..., x_N\}$ with $N$ observations of an r.v. $X \in \mathcal{R}^D$, and that the dataset has been centered such that each dimension has zero mean. 
Furthermore, assume that we'd like to learn an invertible mapping $f: \mathcal{R}^D \rightarrow \mathcal{R}^D$ such that $Y = f(X)$ is distributed according to a simple isotropic Gaussian, i.e. $p_Y(y) = \mathcal{N}(y; 0, I)$. 
In Section \ref{sec:experiments}, we show that distributions with non-Gaussian latent structure can be well-modeled using Gaussian targets and deep architectures.

\subsection{Linear flows (PCA)}
\label{sec:PCA}
If we restrict our functional mapping $f$ to be linear, e.g. $f(X) = WX$ for some matrix $W \in \mathcal{R}^{D \times D}$, then the NF objective of Eq. \ref{eq:nice_objective} reconstructs Principal Component Analysis (PCA). In the case of linear $f$, the objective is written as
\begin{align*}
    W^* &= \argmax_W \frac{1}{N} \sum_{n=1}^N
    \text{log }p_Y(Wx_n) + \text{log } \big| \text{det } W \big|.
\end{align*}
Noting that $\text{log }p_Y(Wx_n) = -\frac{1}{2}|Wx_n|_2^2 + const.$, and that $\text{log } \big| \text{det } W \big| = \frac{1}{2}\text{logdet}\big( W^TW \big)$, this objective becomes
\begin{align}
    W^* &= \argmax_W \frac{1}{N} \sum_{n=1}^N 
    -\frac{1}{2} \norm{Wx_n}_2^2 + \frac{1}{2}\text{logdet}\big( W^TW \big) \nonumber \\
    &= \argmin_W \frac{1}{N} \sum_{n=1}^N 
    \norm{Wx_n}_2^2 - \text{logdet}\big( W^TW \big).
    \label{eq:linear_NF}
\end{align}
Some algebra shows that this objective is equivalent to maximum likelihood estimation of $p_X(x)$ under the assumption that the distribution is Gaussian, i.e. that $p_X(x) = \mathcal{N}(x;0,\Sigma)$. The average log-likelihood of observation set $\dataset{X}$ given parameter $\Sigma$ is written as
\begin{align*}
    L\big(\Sigma; \mathbf{X} \big) &= \frac{1}{N} \sum_{n=1}^N 
    -\frac{1}{2} x_n^T \Sigma^{-1} x_n -\frac{1}{2} \text{logdet}(\Sigma) -\frac{D}{2}\text{log}(2\pi).
\end{align*}
A Gaussian precision matrix is positive semi-definite; therefore, we can write our precision estimate as $\hat{\Sigma}^{-1} = W^T W$ for square matrix $W$, and the MLE objective becomes
\begin{align}
    W^* &= \argmin_W \frac{1}{N} \sum_{n=1}^N
    \frac{1}{2} x_n^T W^TW x_n + \frac{1}{2}\text{logdet}\big( (W^TW)^{-1} \big) \nonumber \\
    &= \argmin_W \frac{1}{N} \sum_{n=1}^N
    \norm{Wx_n}_2^2 - \text{logdet}\big( W^TW \big),
    \label{eq:gaussian_mle}
\end{align}
reproducing the NF objective of Eq. \ref{eq:linear_NF}. 
Given solution $W^*$, the singular value decomposition (s.v.d.) $W^* = USV^T$ yields an orthogonal matrix $V$ whose column vectors $v_i$ are the principal components of the data and a diagonal matrix $S$ whose squared diagonal entries $s_i^2$ are the corresponding precision values (inverse variances).\footnote{The precision estimate is written $\hat{\Sigma}^{-1} = W^T W = (USV^T)^T USV^T = VS^TU^TUSV^T = VS^2V^T $.}

\begin{figure}[t]
    \centering
    \includegraphics[width=\hsize]{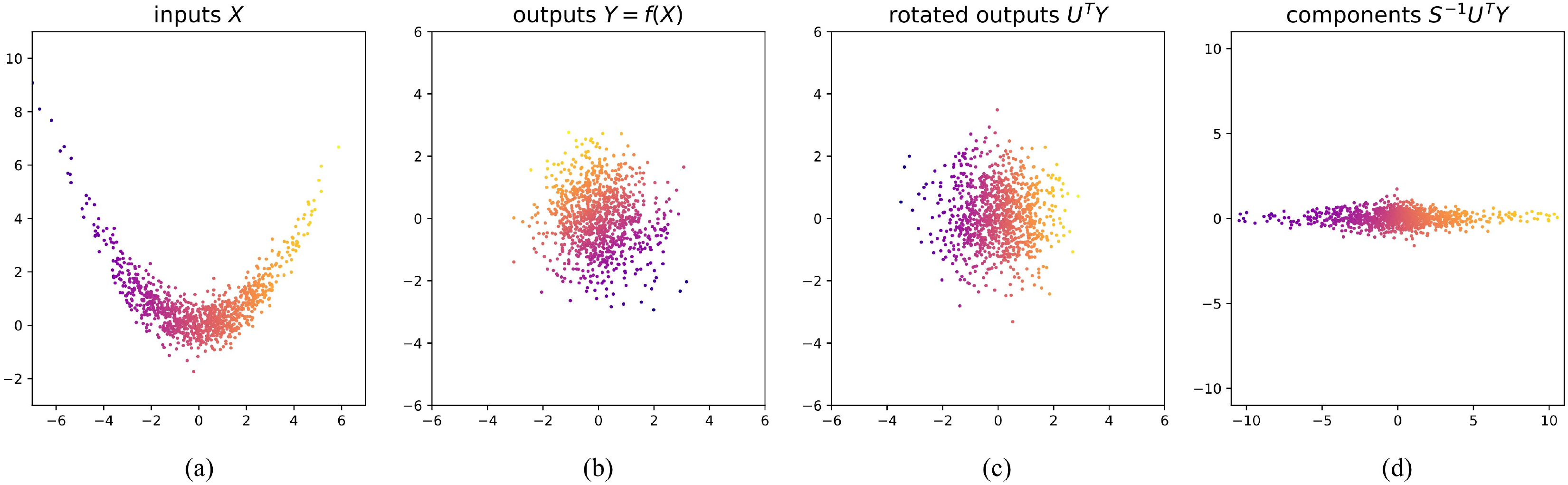}
    \caption{Computing component representations for nonlinear distributions using a function $f$ that maps to isotropic Gaussian. (a) Samples from a 2D nonlinear banana distribution. (b) Transformed samples output by a function $Y = f(X)$ that maps the original distribution to a standard normal. (c) Each output is rotated by a unique orthogonal matrix $U$ determined from the Jacobian s.v.d. at the corresponding input ($J_f(x) = USV^T$). (d) Each rotated output is next scaled by a unique diagonal matrix $S^{-1}$ determined from the Jacobian s.v.d.}
    \label{fig:extracting_components}
\end{figure}

\subsection{Nonlinear flows}
Nonlinear flows provide an extension of the linear case for situations where the data distribution $p_X(x)$ is non-Gaussian. 
For such a distribution, the mapping $Y = f(X)$ must be nonlinear in order to obtain isotropic Gaussian components $Y \sim \mathcal{N}(0,I)$.

For an isotropic Gaussian target distribution, the generalized NF objective of Eq. \ref{eq:nice_objective} is written as
\begin{align}
    f^* = 
    \argmin_f \frac{1}{N} \sum_{n=1}^N 
    \norm{f(x_n)}_2^2 - \text{logdet}\big( J_f(x_n)^T J_f(x_n) \big),
    \label{eq:specific_NF_obj}
\end{align}
where $J_f(x_n)$ is the Jacobian matrix of $f$ evaluated at $x_n$. 
Comparing to the Gaussian MLE objective of Eq. \ref{eq:gaussian_mle}, note that the precision estimate $\hat{\Sigma}^{-1} = W^TW$ has been replaced by an estimate $\hat{\Sigma}^{-1}(x) = J_f(x)^T J_f(x)$ that varies as a function of $x$. The second term of the objective's summation, which can be brought outside of the sum in Eq. \ref{eq:gaussian_mle} if scaled by a constant, must now remain inside the sum, for it depends on $x_n$.

The connection to PCA provides key insights about the nonlinear case. 
Whereas PCA produces a linear whitening transformation for Gaussian data, nonlinear flows produce a \textit{nonlinear whitening transformation} that maps a non-Gaussian variable $X$ to isotropic Gaussian.
As visualized in Fig. \ref{fig:linear_nonlinear_CA}, this nonlinear transformation bears information about the local covariance of $X$ at each region of the input space.
Given the solution $f^*$ that maps $X$ to isotropic Gaussian, the local component directions and variances of the data manifold at each input point are specified by the s.v.d. of the Jacobian matrix, $J_f(x) = USV^T$: the orthogonal matrix $V$ holds the directions in its column vectors, and the diagonal matrix $S$ holds the corresponding inverse standard deviations in its diagonal entries ($\lambda_i = s_i^{-2}$ gives the variance for the $i$-th component).

\section{Un-whitened Latent Components}
\label{sec:projections}
PCA yields a linear basis that serves as a powerful feature representation for Gaussian-like variables. 
Normalizing Flows, which approximate nonlinear component estimation, offer a useful analogue to PCA for non-Gaussian data; however, applications of NFs have focused primarily on sampling and density estimation. 
In some cases, whitened components $Y = f(X)$ are applied as a feature representation for tasks such as manifold traversal and classification \citep{Dinh2017, Nalisnick2019}. 
This feature space is problematic, however, because components that carry little information about the data manifold have been expanded to unit variance.
Here, we describe a simple procedure to extract unwhitened component representations from nonlinaer flows post-training.

Given a non-Gaussian r.v. $X$ and a function $f$ that maps the variable to isotropic Gaussian, the Jacobian $J_f$ can be used to uncover the component projection of each input point. The algorithm consists of inverting the whitening process of $f$ at each output $y_n = f(x_n)$ using the s.v.d. of $J_f(x_n)$. Assuming the Jacobian at $x_n$ factors as $J_f(x_n) = USV^T$, the component projection $\hat{y}_n$ for the point is obtained by transforming the function output by the formula $\hat{y}_n = S^{-1}U^Ty_n$. The first transformation, $U^T$, rotates each $y_n$ such that its dimensions correspond to component axes. 
This step is necessary due to the non-uniqueness of $f$: there may be multiple functions that map the variable $X$ to an isotropic Gaussian. 
The second transformation, $S^{-1}$, scales the components according to their standard deviations, giving emphasis to the components of greatest significance. To ensure consistent output from s.v.d., the columns of $U$ are adjusted such that the loadings with largest magnitude are always positive. 

As a demonstration, consider r.v. $X \in \mathcal{R}^2$ drawn from a banana-shaped distribution, samples of which are shown in Fig. \ref{fig:extracting_components}a. 
The variable is defined as 
\begin{equation*}
X = 
\begin{bmatrix} 
2\epsilon_1  \\
\frac{4}{5}\epsilon_1^2 + \frac{1}{2}\epsilon_2
\end{bmatrix},
\quad \epsilon_i \sim \mathcal{N}(0,1).
\end{equation*}
Furthermore, consider the transformation
\begin{equation*}
f(X) = 
\begin{bmatrix} 
-\frac{1}{4}X_1 - \sqrt{3}X_2 + \frac{\sqrt{3}}{5}X_1^2  \\
\frac{\sqrt{3}}{4}X_1 - X_2 + \frac{1}{5}X_1^2
\end{bmatrix},
\end{equation*}
one of many that maps $X$ to $\mathcal{N}(0, I_2)$.
Samples of the transformed variable $Y = f(X)$ are shown in Fig. \ref{fig:extracting_components}b. 
The Jacobian $J_f$ of this transformation is defined as
\begin{equation*}
\begin{bmatrix}
\frac{\partial f_1}{\partial X_1} & \frac{\partial f_1}{\partial X_2} \\
\frac{\partial f_2}{\partial X_1} & \frac{\partial f_2}{\partial X_2}
\end{bmatrix}
=
\begin{bmatrix} 
-\frac{1}{4} + \frac{2\sqrt{3}}{5}X_1    &  -\sqrt{3} \\
\frac{\sqrt{3}}{4} + \frac{2}{5}X_1     & -1 
\end{bmatrix}
\end{equation*}
and varies as a function of the input. Computing the s.v.d. $J_f(x) = USV^T$ for each point, we first use the left Jacobian singular matrix to rotate each output $y = U^Ty$ (Fig. \ref{fig:extracting_components}c). Next, we use the inverse singular value matrix of each point to scale the dimensions $y=S^{-1}y$ (Fig. \ref{fig:extracting_components}d).
The result is a factorized distribution whose dimensions correspond to the nonlinear components of the data and are scaled according to component variance.

\section{Stability Analysis \& Regularization}
\label{sec:tik_reg}
When the intrinsic dimensionality of the data is less than the apparent dimensionality, the NF optimization algorithm becomes unstable. 
As a simple linear demonstration, consider a 1D Gaussian that is embedded in a 2D space, specified by $p_X(x)$ = $N(x; 0, \Sigma)$. 
The covariance matrix factors as $\Sigma = V \Lambda V^T$, where $\Lambda$ is a diagonal matrix with diagonal entries $[\lambda_1, 0]$. 
The NF algorithm, which is agnostic to the intrinsic dimensionality of the data a priori, attempts to learn a linear mapping $Y=WX$, with $W \in \mathcal{R}^{2 \times 2}$, that maps $X$ to an isotropic Gaussian r.v. $Y \sim N(0, I_2)$. 
Given an observation set $\dataset{X}$, the mapping matrix $W$ is initialized at random with initial singular values $[s_1, s_2]$. 
Recall from Section \ref{sec:PCA} that the optimal mapping $W^*$ has the following property: the inverse squared singular values $s_i^{-2}$ correspond to the eigenvalues of the MLE covariance estimate. 
This covariance estimate is singular; therefore, the optimizer must push one of its singular values to infinity to reach the correct solution. 

In classical statistics, a common technique to mediate singular covariance is to add a small value $\alpha$ to the diagonal of the empirical covariance $S$, such that $S' = S + \alpha I$. 
This technique, known as Tikhonov regularization (or \textit{shrinkage}), provides additional robustness to statistical estimators at the cost of a small, often negligible estimation error. 
To understand the affect of this modification on Gaussian MLE, consider again the MLE objective of Eq. \ref{eq:gaussian_mle}. 
Given empirical covariance $S = \frac{1}{N} \sum_{n=1}^N x_n x_n^T$, the objective can be re-written as
\begin{align*}
    W^* &= \argmin_W
    \text{ tr}(S W^TW) - \text{logdet}\big( W^TW \big).
\end{align*}
Adding a diagonal to the empirical covariance, this objective becomes
\begin{align*}
    W^* &= \argmin_W 
    \text{ tr}((S + \alpha I) W^TW) - \text{logdet}\big( W^TW \big) \\
    &= \argmin_W 
    \text{ tr}(S W^TW) - \text{logdet}\big( W^TW \big) + \alpha * \norm{W}_F^2.
\end{align*}
Thus, Tikhonov regularization is implemented by simply adding an L2 penalty on $W$ to our optimization objective, with penalty weight $\alpha$.
In the linear case, the Jacobian of the function $f$ that maps $X$ to an isotropic Gaussian is specified by the matrix $W$ and is independent of the input. In general, however, the Jacobian matrix $J_f(x)$ is a function of $x$. Generalizing Tikhonov regularization to nonlinear component estimation, we place an L2 penalty on the Jacobian evaluated at each input. The objective of Eq. \ref{eq:specific_NF_obj} then becomes
\begin{align*}
    f^* &= \argmin_f \frac{1}{N} \sum_{n=1}^N 
    \norm{f(x_n)}_2^2 - \text{logdet}\big( J_f(x_n)^T J_f(x_n) \big) + \alpha*\norm{J_f(x_n)}_F^2.
\end{align*}
Thus, applying Tikhonov regularization to Normalizing Flows is as simple as placing an L2 penalty on the Jacobian matrix $J_f(x_n)$ at each data point $x_n$. 
This regularization term stabilizes the algorithm and facilitates accurate nonlinear component estimation with black-box optimization (see Appendix A for simple demos).

\section{Experiments}
\label{sec:experiments}
\begin{figure}[t]
    \centering
    \includegraphics[width=\hsize]{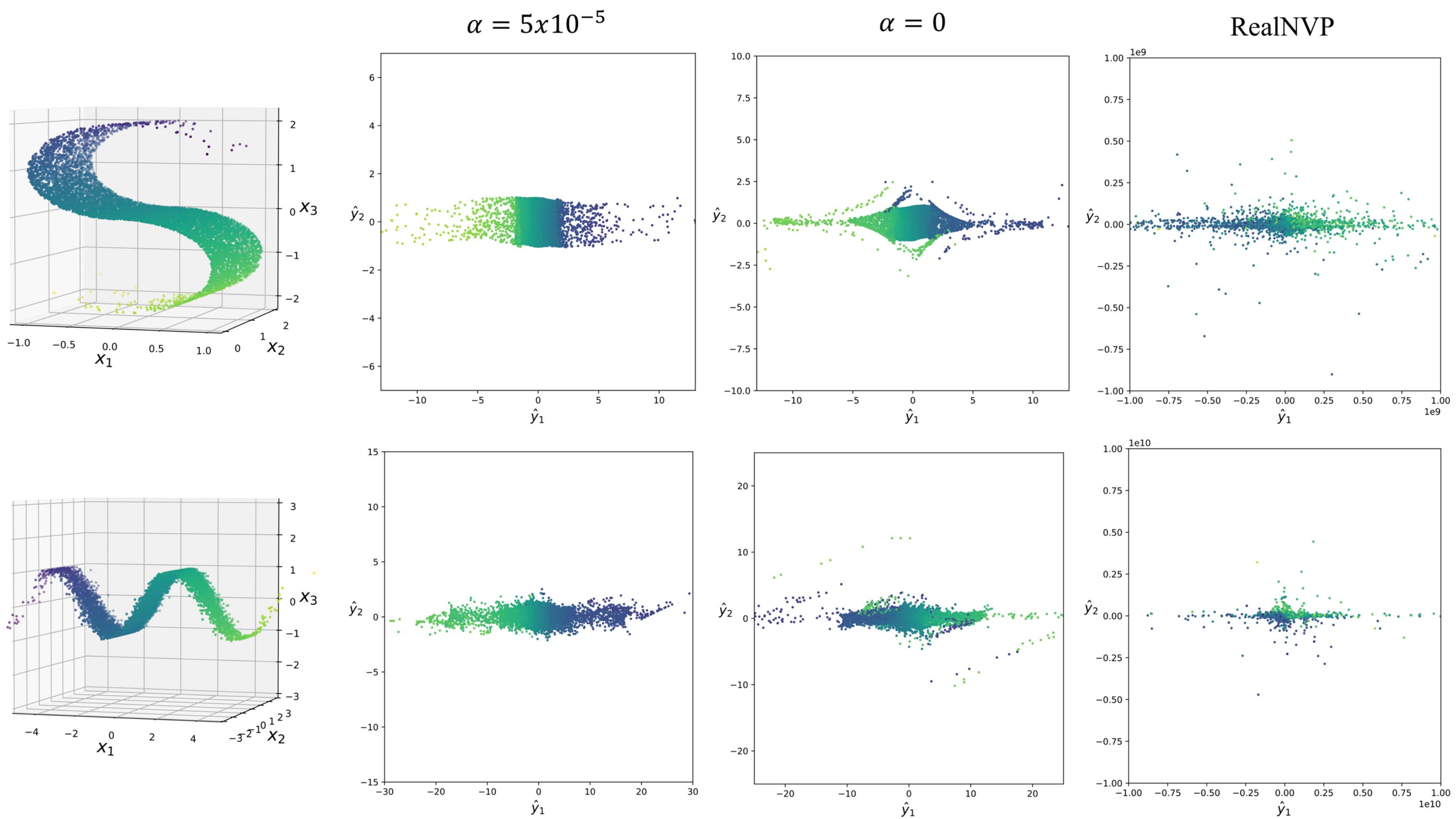}
    \caption{Results of Normalizing Flows applied to two nonlinear manifolds. The source distributions (left) are 2D manifolds embedded in a 3D space (top row: ``S-curve" dataset, bottom row: sine wave dataset). Plots show the nonlinear component projections of the data computed with each model after training. Our regularized model ($\alpha=5\times10^{-5}$) succesfully estimates the nonlinear components. Neither an unregularized version of our model ($\alpha = 0$) nor an unmodified RealNVP architecture produce consistent component projections.}
    \label{fig:3d_dists}
\end{figure}

We asked whether Normalizing Flows could produce consistent component estimates when trained with Tikhonov regularization and applied with our feature extraction algorithm, focusing on datasets with reduced intrinsic dimensionality.
To provide a controlled setting where the effects of our tools could be isolated, we applied generalized invertible architectures with minimal restriction (Appendix B).
For an input variable with D dimensions, we constructed a multilayer feed-forward network composed of L hidden layers, each consisting of an $\mathcal{R}^D \rightarrow \mathcal{R}^D$ affine transformation coupled with an invertible, element-wise nonlinearity.
An output layer with linear activation was appended to the end of the network. 
For each dataset and architecture, the network was initialized at random and trained via mini-batch gradient descent with a batch size of 200 and the Adam optimizer. 
We used an isotropic Gaussian target distribution $p_Y(y) = \mathcal{N}(y; 0, I)$, with $Y \in \mathcal{R}^D$. 
Training data were centered such that each feature has zero mean.
The experiments described here can be replicated using the provided code (Appendix C).

\subsection{Nonlinear manifold learning}
We began with two manifold learning tasks motivated by the nonlinear dimensionality reduction literature \citep{Xing2016}, shown in in Fig \ref{fig:3d_dists}. 
Each dataset consists of 10,000 observations from a 2D latent distribution that has been embedded in a 3D space by a nonlinear transformation. 
The first dataset is a version of the synthetic ``S-curve” manifold with one of two latent factors distributed as Gaussian. 
The second dataset consists of a sine wave with 2D Gaussian latent structure ($X = [2\epsilon_1, \epsilon_2/2, \text{sin}(2\epsilon_1)$], $\epsilon_i \sim \mathcal{N}(0,1)$).
For each of these datasets, we applied a multilayer architecture with 8 hidden layers and the inverse hyperbolic sine nonlinearity. 
This nonlinearity has the nice property of a smooth gradient and both a domain and range of $\mathcal{R}$. 
With Tikhonov regularization applied ($\alpha = 5 \times 10^{-5}$), we found that the network learns a strong estimate of the underlying 2D latent structure in both cases. 
Plots show the projections $\hat{Y}$ of each point onto the top 2 components of the learned model, obtained using the algorithm from Section \ref{sec:projections}. 
Interestingly, the network is able to uncover non-Gaussian latent structure from the S-curve dataset using Gaussian component targets. 
Without regularization, the network fails to produce a coherent estimate of the underlying latent structure in either case.

\begin{figure}[t]
    \centering
    \includegraphics[width=0.8\hsize]{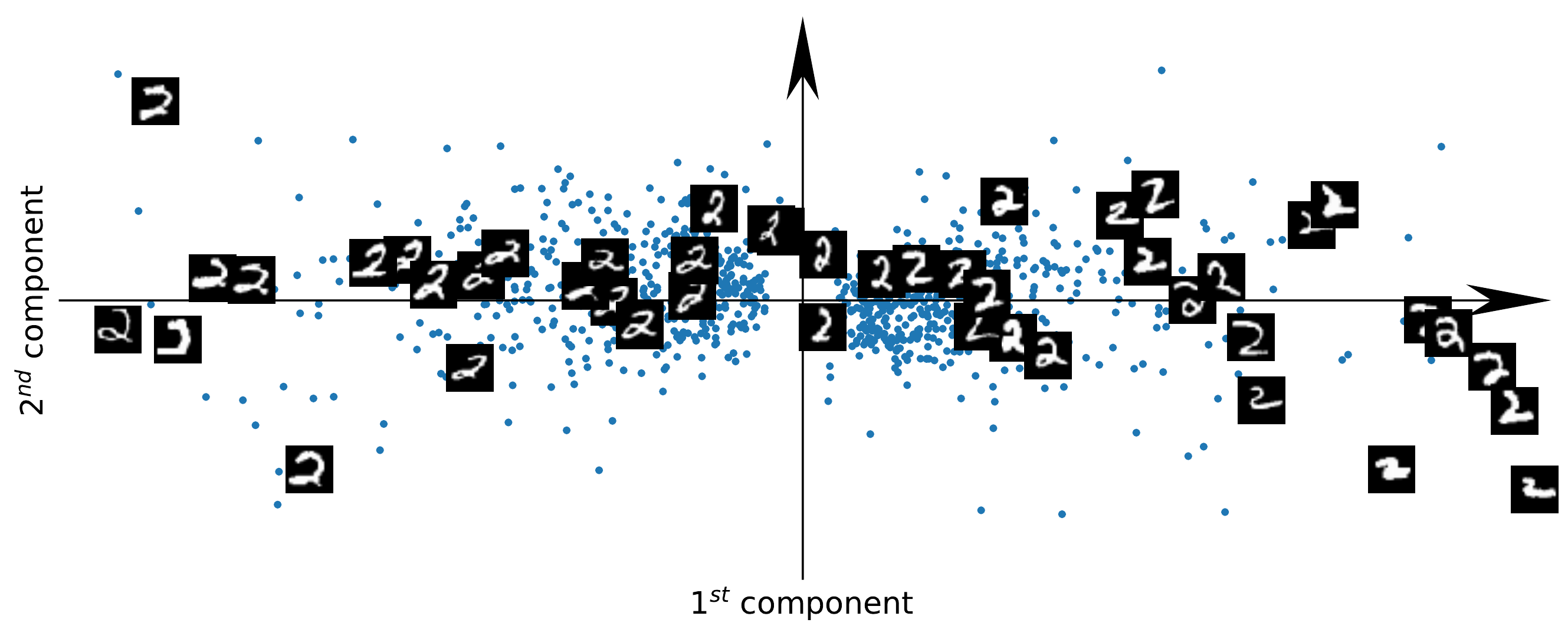}
    \caption{Top 2 component projections for MNIST 2's. Projections for 1,000 training images are obtained from our regularized MNIST model ($\alpha = 5 \times 10^{-5}$) using the algorithm of Section \ref{sec:projections}.}
    \label{fig:mnist_components}
\end{figure}

We further tested the RealNVP architecture (Appendix B.2) on these same two nonlinear distributions, again using Adam with a batch size of 200. RealNVP is an invertible feed-forward architecture composed of simple coupling layers with triangular Jacobian, lending to cheap likelihood computation. 
We used a model with 6 coupling layers, each composed of log-scale and shift functions parameterized by neural networks with 2 hidden layers of 512 ReLU units (defaults of the TensorFlow RealNVP implementation), and a masking dimensionality $d=1$. Following standard practices, the dimensions are permuted after each layer and batch normalization is applied to improve training performance.
Interestingly, the network's training loss converges without Jacobian regularization\footnote{We note that, when applied to the MNIST dataset, our unregularized RealNVP architecture does not converge. We hypothesize that, in our simple 3D tasks, batch normalization may be sufficient for convergence. With higher-dimensional data, where the difference between the apparent and intrinsic dimensionalities is large, a more theoretically-sound regularization technique is required.}, and samples from the model are visually consistent with samples from both our regularized model and ground truth. 
However, component projections extracted from this architecture lack coherent structure and are many orders of magnitude larger than those of the standard network.

\subsection{Learning MNIST digits}
Our 3D manifold learning tasks demonstrate that regularized flows can learn a component representation that is both disentangled and scaled when presented with simple nonlinear distributions. 
We next asked whether our framework could produce analogous representations for more complex data with higher dimensionality. 
The MNIST dataset of handwritten digits provides a setting of increased complexity over our 3D distributions where data points are easy to visualize and interpret.
In our second experiment, we tested whether a regularized flow could produce coherent low-dimensional component representations of MNIST images.
The dataset consists 28x28x1 images with pixel values in range [0,1]. 
We trained our model on the manifold of image class ``2”, providing 5,958 images for training and 1,032 for validation.
Images were flattened to dimensionality 784, and we applied an architecture from the family discussed above with 6 hidden layers and inverse hyperbolic sine nonlinearity. 
The model was trained with regularization weight $\alpha=5 \times 10^{-5}$.
After training, we selected 1,000 images at random from the training set and used our component projection algorithm to extract the top 2 components of the model for each image (Fig. \ref{fig:mnist_components}). 
The model produces a qualitatively meaningful representation of the image manifold wherein 2's of similar drawing style are grouped together, capturing a combination of properties such as digit angle, line thickness, and curved vs. straight lines.

\begin{figure}[t]
    \centering
    \includegraphics[width=\textwidth]{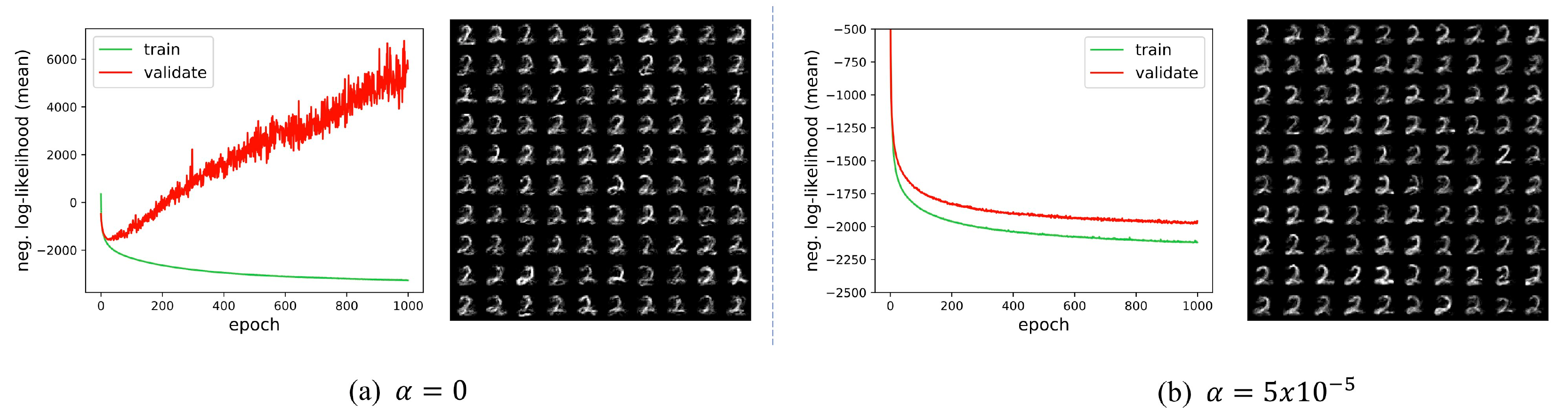}
    \caption{
    Regularized vs. unregularized Normalizing Flows applied to MNIST 2's.
    (a) With no regularization, the model overfits severely to the training data. 
    (b) With regularization applied, the model exhibits robust generalization performance. 
    Samples from the resulting model are shown next to each plot.
    }
    \label{fig:mnist_overfitting}
\end{figure}

In one final experiment, we set out to test the effect of Tikhonov regularization on likelihood generalization and image sample quality. 
A primary question of ours was whether the quality of generated samples alone is sufficient proof of a generalized density model. 
We re-trained our MNIST flow from above with no regularization ($\alpha=0$), keeping all other hyperparameters fixed.
Learning trajectories on the train and validation sets, as well as samples from the resulting model, are compared for our regularized and unregularized models in Fig. \ref{fig:mnist_overfitting}.
The unregularized MNIST model exhibits significant over-fitting during the course of training, achieving average log-likelihoods of 3,265 and -5,659 on the training and validation sets, respectively. 
In contrast, the Tikhonov-regularized model exhibits strong generalization performance, achieving train and validation scores of 2,111 and 2,014. 
The regularized model has notably better sample quality overall, characterized by less noise in many of the images. 
Interestingly, samples generated by the unregularized model share many consistencies with ground truth, in spite of poor generalization performance on the log-likelihood task. 
This result suggests that sample quality alone may provide insufficient proof of a robust generative model.
One possible explanation is that the unregularized model learns to memorize the training data, whereas the regularized model learns a generalizable feature representation. 
In future work, we'd like to test this hypothesis more rigorously with additional experiments.

\section{Discussion}
\label{sec:discussion}
In this work, we provide a new theoretical framework for understanding and analyzing Normalizing Flow algorithms.
We developed a component extraction method for NFs that allows for new interpretation of the specific basis features in a manner analogous to linear methods such as PCA. 
Using this new method, we found that NF models indeed learn a meaningful low-dimensional representation of the input.
In future work, we'd like to test these representations on downstream tasks such as classification and manifold traversal. 
In addition, we'd like to apply our feature extraction method to analayze and understand the representations of other algorithms that learn a mapping between isotropic Gaussians and complex distributions. 
These include (but are not limited to) the generator network of a GAN and the decoder of a VAE.

A key insight of our study is the need for Tikhonov regularization in NF models.
Tikhonov regularization has been applied in many different neural network contexts \citep[e.g.,][]{Bishop1995, Woo2005, Rifai2011}; however, the interpretation of regularization, and its implications for algorithmic performance/stability, varies greatly based on the specific objective function, architecture and algorithm. 
Our work provides an original interpretation of Tikhonov regularization in the context of NF models. By drawing a connection to local covariance shrinkage, we highlight a major stability concern for the NF objective and provide a theoretically-grounded solution to mediate it.
\citet{Bishop1995} addresses supervised training of neural networks for regression/classification, and thus bears little implication for the unsupervised learning of probability distributions. 
\citet{Rifai2011} demonstrate a Jacobian-based regularization penalty for neural nets, here in the context of unsupervised learning with autoencoders. 
Although their L2 penalty appears similar to our own, autoencoders have no direct connection to fundamental component analysis in the way that Normalizing Flows do. 
A linear autoencoder learns the subspace spanned by the top principal directions, but not the directions themselves \cite{Kunin2019}. 
In contrast, a linear flow directly learns principal directions (Sec. \ref{sec:PCA}). 
Importantly, the need for our regularization penalty is attributed to the fact that we are learning principal directions and estimating eigenvalues of the MLE covariance (Sec. \ref{sec:tik_reg}). 
Finally, \citet{Woo2005} demonstrate Tikhonov regularization of a nonlinear ICA model. 
Although the problem is loosely related to ours, their mutual information objective is quite different from NFs and specific to the source separation context. 
Further, their ad-hoc regularization term is difficult to interpret, and they provide little motivation beyond parameter constraints.

\bibliographystyle{plainnat}
\bibliography{references}

\newpage
\appendix
\section{Regularization Demo}
\label{sec:tikonov_demo}
\begin{figure*}[ht]
    \centering
    \subfigure[Linear flow demo]{
        \centering
        \includegraphics[width=\hsize]{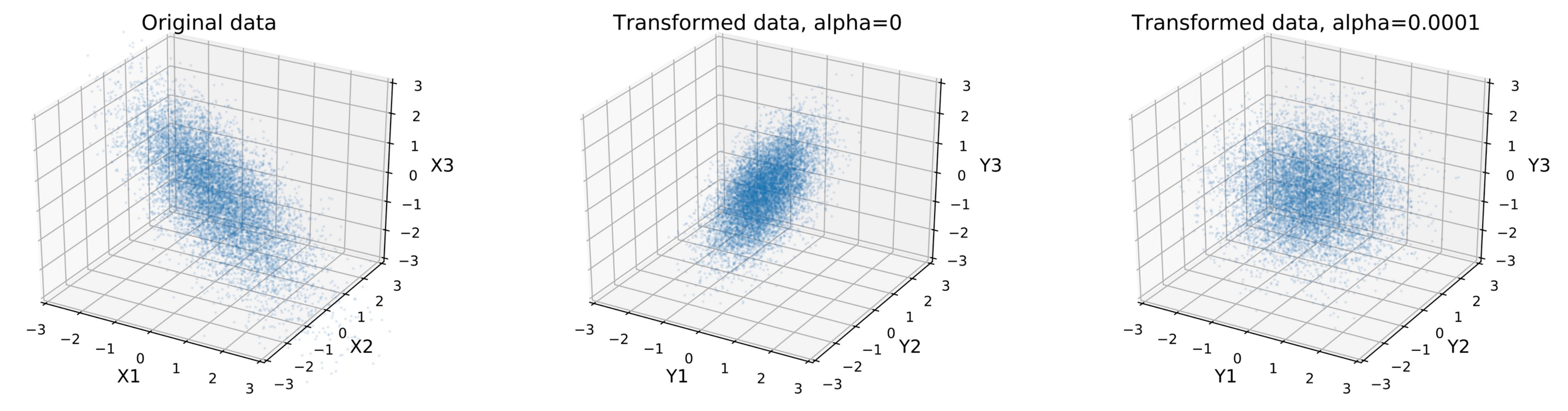}
    }
    \subfigure[Nonlinear flow demo]{
        \centering
        \includegraphics[width=\hsize]{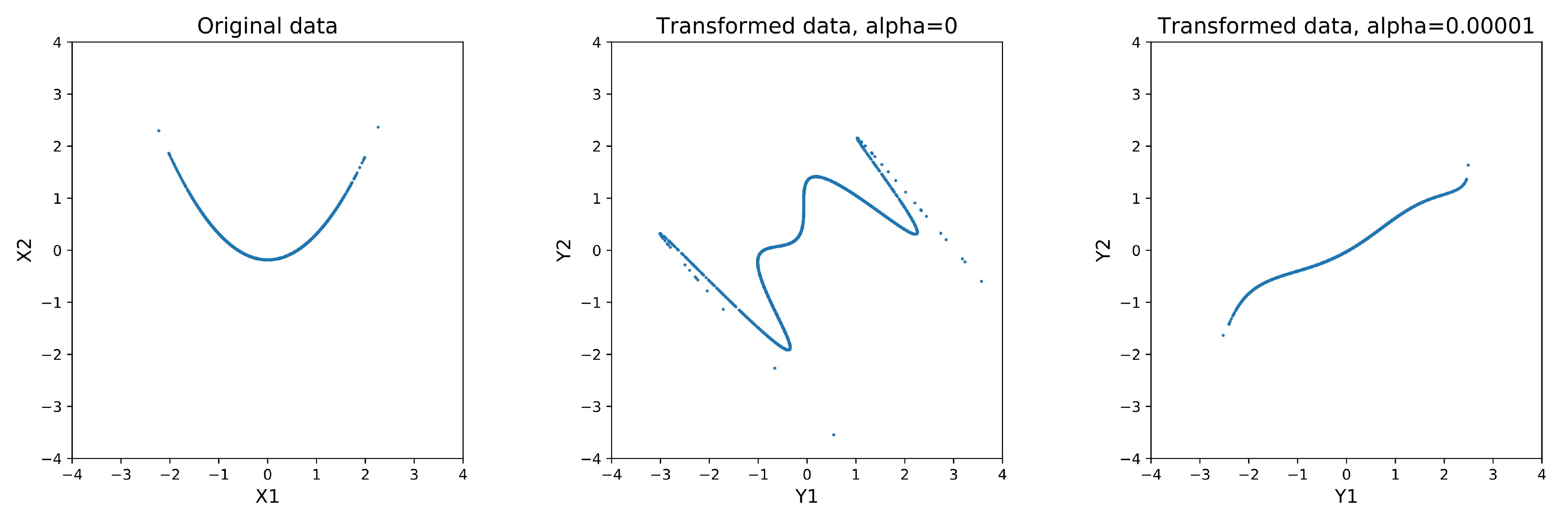}
    }
    \caption{Tikhonov regularization stabilizes Normalizing Flow training. (a) A linear flow was applied to observations from a 2D Gaussian embedded in a 3D space. The original centered dataset is shown (left). The transformation $Y = WX$ was initialized at random and optimized via gradient descent. Without Tikhonov regularization, the algorithm fails to converge to the correct components (middle). With regularization weight $\alpha=10^{-4}$, the algorithm converges to the correct data components (right). (b) A nonlinear flow was applied to observations from a 1D nonlinear data manifold embedded in a 2D space. The original centered dataset is shown (left). The function estimator $Y = f(X)$ is parameterized by a neural network with 6 hidden layers of 2 units, each with softplus nonlinearity, and a final linear layer with 2 units. The network parameters are initialized at random and optimized via gradient descent. Without Tikhonov regularization, the algorithm fails to converge to the correct components (middle). With regularization weight $\alpha=10^{-5}$, the algorithm converges to the correct data components (right).}
    \label{fig:regularization_demo}
\end{figure*}

Fig. \ref{fig:regularization_demo} shows a demonstration of the affect of Tikhonov regularization in two simple Normalizing Flow (NF) use cases, one linear and another nonlinear. 
In each case, the observed variable has an intrinsic dimensionality that is smaller than the number of inputs. 
The optimal mapping $f$ is determined by gradient-based optimization. 
For the linear case, the mapping is specified by $Y = WX$, where $W$ is a weight matrix that is learned during optimization. 
For the nonlinear case, the function is specified as $Y = f(X)$, where $f$ is parameterized by a simple invertible neural network model whose parameters are learned during optimization. 
In each case, without regularization the optimization algorithm fails to converge to the correct solution. 
When Tikhonov regularization is added to the objective, the algorithm successfully converges to the correct solution.

\section{Generalized Invertible Function Approximators}
\label{sec:general_nice_architecture}
In our experiments, we use a generalized deep neural network architecture to parameterize the NF invertible function approximator. 
This contrasts with recent works from the NF literature, which have used restricted architectures to make Jacobian determinant computations more tractable (see \ref{sec:normflow_architecture} and \ref{sec:realnvp_architecture}). 
Our experiments are designed to isolate the affect of Tikhonov regularization on the NF objective. 
Although our algorithm is agnostic to choice of architecture, and therefore can be applied to restricted models as well, we use a generalized architecture in our experiments and sideline concerns about training speed. 
For an observed variable $X \in \mathcal{R}^D$, our neural network is composed by stacking $L$ standard layers, each with an affine transformation followed by an element-wise non-linearity. Each transformation is $f: \mathcal{R}^D \rightarrow \mathcal{R}^D$.
Here, we briefly review a few restricted architectures from the Normalizing Flows literature.

\subsection{Planar \& Radial Flows}
\label{sec:normflow_architecture}
\citet{Rezende2015} introduced a specialized class of architectures composed of nonlinear transformations called \textit{flow layers}. 
These layers are designed to exploit various determinant identities, such as the Matrix Determinant Lemma, and facilitate fast Jacobian determinant computation. 
The original paper proposed two types of flow layers: \textit{planar flows} and \textit{radial flows}.

\subsection{RealNVP}
\label{sec:realnvp_architecture}
RealNVP \citep{Dinh2017} is a specialized architecture composed of partitioned nonlinear transformations called \textit{coupling layers}. Coupling layers are designed to have triangular Jacobian, making Jacobian determinant computation very fast.
For each layer, the user specifies a partitioning dimensionality $d$, and the transformation that maps inputs $X \in \mathcal{R}^D$ to outputs $Y \in \mathcal{R}^D$ is defined as
\begin{equation}
    \label{eq:realnvp}
    \begin{aligned}
    Y_{1:d} &= X_{1:d} \\
    Y_{d+1:D} &= X_{d+1:D} \odot \text{exp}\big( s(X_{1:d}) \big) + t(X_{1:d}),
    \end{aligned}
\end{equation}
where $s$ and $t$ are the log-scale and shift functions, respectively.
RealNVP is an extension of an earlier model that used coupling layers without a log-scale function \citep{Dinh2015}.


\section{Code}
\label{sec:code}
All experiments from this paper can be replicated using the source code repository located at:
\begin{center}
    \textbf{\texttt{https://github.com/zdf233/normflow-theory}}
\end{center}
Instructions for running the experiments are provided in the \texttt{README.md} file.

\end{document}